\documentclass[runningheads]{llncs}
\usepackage[T1]{fontenc}
\usepackage{graphicx}
\usepackage{xcolor}
\usepackage{footmisc}

\begin{document}

\title{How LLMs Fail to Support Fact-Checking}

\author{Adiba Mahbub Proma$^{*}$ \and 
        Neeley Pate$^{*}$ \and
        James N. Druckman \and
        Gourab Ghoshal \and
        Hangfeng He \and
        Ehsan Hoque}% <-this % stops a space
\institute{University of Rochester}

% \author{
% \IEEEauthorblockN{Adiba Mahbub Proma\IEEEauthorrefmark{1}, Neeley Pate\IEEEauthorrefmark{1}, 
% James Druckman, Gourab Ghoshal, Hangfeng He, and Ehsan Hoque}
% \IEEEauthorblockA{University of Rochester \\
% Email: \{aproma, npate, mehoque\}@cs.rochester.edu, 
% jdruckma@UR.Rochester.edu, gghoshal@pas.rochester.edu, hangfeng.he@rochester.edu}
% \thanks{\IEEEauthorrefmark{1}Equal contribution.}
% }

\maketitle

\textit{* denotes equal contribution}  
\begin{abstract}
Large language models (LLMs) have been shown to both create misinformation, often trusted by end users, and combat misinformation when supported by specialized pipelines. Given users may prompt LLMs to fact-check information without access to such pipelines, it is crucial to understand how base models perform fact-checking. In this paper, we empirically study the capabilities of three LLMs – ChatGPT 4o-mini, Gemini 1.5-flash, and Claude 3.5-haiku – in countering misinformation and supporting factual information. We implement a two-step prompting approach, where models first identify credible sources for a given claim and then generate a summarized argument. Our findings suggest that models struggle to ground their responses in real news sources, and tend to prefer citing left-leaning sources. This work highlights concerns about using LLMs for fact-checking through only prompting, emphasizing the need for more robust guardrails on the base systems and bringing awareness of these pitfalls to end users. Our findings have implications for misinformation and bias dissemination as users turn to LLMs as fact-checkers, and highlight a new challenge in information navigation.
\end{abstract}

% \marklessfootnote{*Equal contribution.} \\
% \marklessfootnote{1 corresponding author: aproma@cs.rochester.edu}

% \begin{abstract}
% Misinformation can invoke various emotional responses among individuals, which may distract users from critically thinking about the information presented to them. While Large Language Models (LLMs) can amplify online misinformation, they also show promise in tackling misinformation. Given users may prompt LLMs for fact-check information, it is crucial to understand model capabilities to do so. In this paper, we empirically study the capabilities of three LLMs – ChatGPT 4o-mini, Gemini 1.5-flash, and Claude 3.5-haiku – in countering misinformation and supporting factual information. We implement a two-step, chain-of-thought prompting approach, where models first identify credible sources for a given claim and then generate a summarized argument. Our findings suggest that models struggle to ground their responses in real news sources, and tend to prefer citing left-leaning sources. This work highlights concerns about using LLMs for fact-checking through only prompting, emphasizing the need for more robust guardrails. \textcolor{red}{Our findings have implications for misinformation and bias dissemination as users turn to LLMs as fact-checkers, and highlights a new challenge of information navigation.}
% \end{abstract}

 \\

\noindent
% \footnote{$^{*}$ Equal contribution.} \\
% \marklessfootnote{$^{1}$  Corresponding author: aproma@cs.rochester.edu}

% FROM THE IDeaS DESCRIPTION
% One consequence of having these discussions on these platforms is that it creates an online breeding ground for growing and disseminating misleading information of varying levels of credibility and the spread of hate speech and extremism. In this conference we ask: How is this done? Who is doing it? Why is it being done? What are the social consequences? How can it be countered?
% We invite papers that address these questions and are particularly interested in papers that touch on the role that non-credible information, hate speech and extremism are playing in events related to health, medicine, elections, conflict, diplomacy, and community resilience. Policy, empirical, qualitative, data science and simulation papers are of equal interest. The conference seeks to be a broad look at online harms: thus issues such as platform regulation, new technologies (e.g. deep fakes), human psychological and social response, links between online and offline behavior, are all relevant.

\vspace{-1em}

\section{Introduction}
%How humans perceive, process and interact with information online is strongly influenced by various cognitive and affective processes. Specifically, 
\vspace{-0.5em}

% potential reframe

% Users increasingly trust LLM responses
% Users increasingly use LLMs for asking questions
% LLMs are known to hallucinate

Large language models (LLMs) have increasingly become a trusted and useful tool to the average end user, shown to be more empathetic than human-written responses \cite{lee2024large}. However, LLMs pose unique threats as fact-checkers, as users tend to place high trust in LLMs, which can increase users' susceptibility to misinformation \cite{deverna2024fact}. While LLMs have been used to combat misinformation through various specialized pipelines, such as utilizing Retrieval-Augmented Generation (RAG) \cite{lewis2020retrieval,promaumap}, end users often only have access to a web-hosted chat window and utilize prompting \cite{knoth2024ai}. Therefore, it is crucial to understand how different LLMs perform in finding supporting or debunking information given a claim, without the support of specialized systems, which is most accessible to non-technical users.

%means such as detection \cite{lewis2020retrieval,lucasetal2023fighting} and

%LLMs even when the LLM provides false information later within interactions \cite{amaro2023believe} or mislabel statement factuality \cite{deverna2024fact}. High levels of trust can also backfire and increase users' susceptibility to misinformation \cite{deverna2024fact}. 

Prior works highlighted various pitfalls of LLMs relevant to fact-checking. Hallucination, or the generation of false information \cite{Augenstein_Baldwin_Cha_Chakraborty_Ciampaglia_Corney_DiResta_Ferrara_Hale_Halevy_etal._2024}, paired with high confidence in their responses, is one way an LLM may mislead a potential user \cite{Augenstein_Baldwin_Cha_Chakraborty_Ciampaglia_Corney_DiResta_Ferrara_Hale_Halevy_etal._2024}. LLMs may fail to cite source material used to generate the response \cite{peskoff-stewart-2023-credible}, creating a roadblock for end users to verify LLM-generated content. While LLMs are constantly advancing, they often come with ``training cut-offs'', meaning the LLM may be ill-equipped to fact-check events after that date \cite{Augenstein_Baldwin_Cha_Chakraborty_Ciampaglia_Corney_DiResta_Ferrara_Hale_Halevy_etal._2024}.

\begin{figure}[ht]
    \centering
    \includegraphics[width=0.8\linewidth]{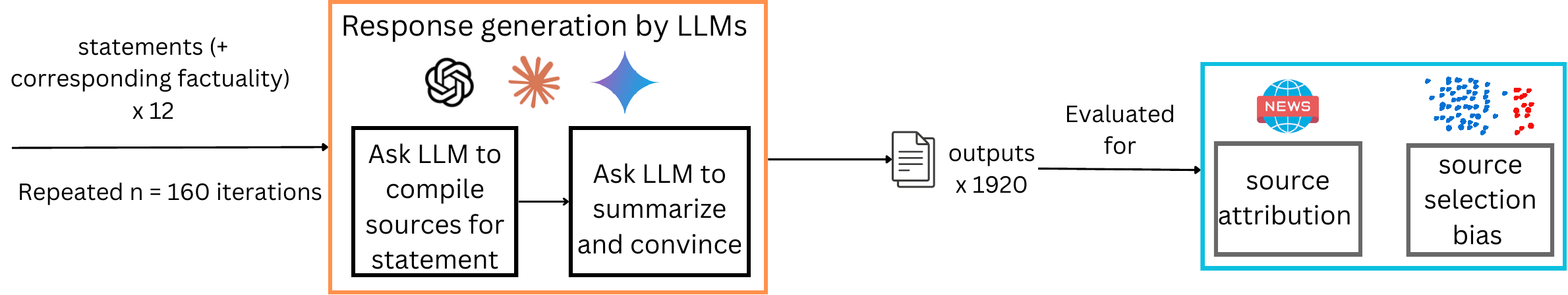}
    \caption{Overview of response generation and evaluation. For 12 statements, 160 iterations are generated, resulting in 1920 data points. The data points are evaluated for source attribution and source selection bias within the output.}
    \label{fig:Baseline_Flow}
\end{figure}

%Methods + Results 
In this paper, we aim to experimentally understand the capabilities of LLMs in fact-checking statements within the political domain. We consider one model from ChatGPT, Gemini, and Claude and utilize a two-step process  where we prompt the models first to select credible sources for a given claim and corresponding factuality (True/False), and then ask the models to summarize the information (Figure \ref{fig:Baseline_Flow}). Analysis of the intermediate and final outputs show that the selected LLMs find it challenging to draw from real, credible information from sources and often select left-leaning sources over others. These findings shed light on the limitations of current LLMs in combating misinformation through source attribution. Our findings have implications for users who use LLMs for fact-verification, including potential exposure to unintentional misinformation and political biases. Empirically evaluating niche aspects of LLM responses, such as within the news domain, helps us pinpoint future scopes for improvement of LLMs.
% tend to prefer left-leaning sources.

% \input{related_works}

\section{Methods}

We utilize a two-step process to generate LLM response: source curation and rhetorical styling. In stage 1 (source curation), the LLMs are asked to prove or disprove the provided statement by finding evidence based on the corresponding factuality (i.e., whether the statement is true or false). We prompt the LLM to identify information from 10 news outlets - 4 left-leaning, 3 center-leaning, and 3 right-leaning, as defined in prior work \cite{promaumap} - balancing representation of political leanings to reflect diverse user preferences. We also prompt the LLMs to provide the corresponding headline to support their arguments. This is later used to verify whether or not these articles exist. The output of stage 1 acts as the input to stage 2 (rhetorical styling), where the LLMs are asked to summarize the facts from stage 1. We use appropriate guardrails to prevent offensive language. This summarization step is crucial as it is intended to reflect the output a typical user might encounter.

We analyze the following two areas: existence of source-generated citations and model source preference. We chose ChatGPT \cite{achiam2023gpt} and Gemini \cite{team2023gemini} families of models since they have been used for misinformation research \cite{huang2024fakegpt,omar2025generating}, and Claude \cite{bai2022constitutional} family of models since Anthropic has taken measures to curb misinformation in their models \cite{anthropic}. Specifically, we use GPT 4o-mini, Gemini 1.5-flash and Claude 3.5-haiku. Crucially, each of these families offers a UI, making them easily accessible to non-technical users.

\subsection{Selection of Statements for Evaluation}
For evaluation, the experiment design utilizes 12 unique statements, following those used in prior works \cite{promaumap}. The statements cover both economic and non-economic topics, and each statement is categorized by factuality (True/False). The response generation for each statement was repeated 160 times to capture any potential variability in outputs, resulting in 1920 data points per model for evaluation.
%\footnote{\url{https://news.gallup.com/poll/651719/economy-important-issue-2024-presidential-vote.aspx}}

%and used to generate responses, repeating 160 times per input statement. 

% \begin{table}[t]
%     \centering
%     \begin{tabular}{p{0.2\linewidth} | p{0.6\linewidth}| p{0.1\linewidth}}
%        \hline
%        \textbf{Statement Category}  &  \textbf{Statement Sample} & \textbf{Factuality} \\
%        \hline
%        Fuel and Oil & Electric vehicles and hybrid vehicles have surged in popularity in the United States, making up 30\% of all cars on the road. & False \\
%        \hline
%        Overpopulation & The provisional number of births for the United States in 2023 was 3,591,328, down 2\% from 2022. & True \\
%        \hline
%        Inflation & Only Americans under 50 are affected by the increased recent inflation. & False \\
%        \hline
%        Immigration & The number of people in the U.S. illegally is upwards of 20, 25, maybe 30 million. & False \\
%        \hline
%     \end{tabular}
%     \caption{Statement categories, with sample statement and associated factuality. Each category has 3 statements, with 2 being false and 1 being true.}
%     \label{statements}
% \end{table}
\vspace{-0.5em}
\subsection{Evaluation of Generated Sources}
\vspace{-0.5em}
In stage 1, LLMs are asked to provide the headline and corresponding news source for their supporting information in order to verify whether the exact or similar article exists. To validate this, we utilize GNews\footnote{\url{https://pypi.org/project/gnews/}}, an API that returns information about news articles, including the news source, headline, and the date of publication. GNews is fed the (source, headline) pair generated by the LLM and returns the top 10 results based on that search. The source is included so that GNews is more likely to return headlines published by the given source, and the headline is provided to match keywords in real headlines. Since we want to consider the real headline most similar to the generated headline, we use cosine similarity\footnote{The cosine similarity scores of two texts can fall within the range of [0.0, 1.0]; 1.0 represents highly similar content (or a ``match''), 0.0 represents high dissimilar content, and 0.5 acts as our cutoff for defining ``similar'' headlines.} to measure text similarity or diversity, as done in prior work \cite{promaumap}. As GNews returns 10 results total, we select the headline with the highest cosine similarity score for final classification. Along with headline similarity, we also consider the publication date. Because input statements and factuality were derived for the 2023-2024 time frame, this constitutes the optimal publication time\footnote{Data was collected and analysis was conducted in December 2024 and January 2025.}. Considering both headline similarity and publication date, we {classify the top result into one of the following: \\

\noindent
\textit{Tier 1}: One of the returned headlines matches the generated headline, falls within time frame.

\noindent
\textit{Tier 2}: One of the returned headlines matches the generated headline, falls outside time frame.

\noindent
\textit{Tier 3}: One of the returned headlines is similar to the generated headline but not exact, falls within time frame.

\noindent
\textit{Tier 4}: One of the returned headlines is similar but not exact to the generated headline, falls outside time frame.

\noindent
\textit{Tier 5}: All of the returned headlines are not similar to the generated headline (all cosine similarity scores fall below 0.5).

\noindent
\textit{Incomplete Entry}: Either the source or the headline is missing, thus a proper search cannot be conducted.

% For determining whether or not the top pair falls within tier 5, we utilize a cosine similarity cutoff of 0.5. Our reason for selecting 0.5 as a cutoff is because cosine similarity ranges from 0.0 - 1.0, and so, 0.5 would be a generous middle ground.

% \begin{figure}[t]
%     \centering
%     \includegraphics[width=0.8\linewidth]{imgs/evaluation_prompts2.pdf}
%     \caption{Prompt used for LLM evaluation. An example input and output also show what the generations may look like.}
%     \label{fig:evaluation_prompts}
% \end{figure}

\subsection{Evaluation of Source Preference}
\label{tierdesc}

\vspace{-0.5em}
% 2) source bias

After summarizing the output from stage 1, we count the frequency of source appearances across all outputs. If none of the 10 sources are found within the final output, the response is categorized as `Other'.

% We quantify the number of final outputs in which each source appears to measure any potential LLM-imposed biases.

%or if the models select sources not within our 10 sources. Based on the final outputs generated by the models, we scan the output for any of the expected 10 sources. 

% \subsection{Measuring Persuasiveness of Final Outputs}
% Given the overall goal of the response generation is to create persuasive arguments while drawing from existing sources, there needs to be some metric to evaluate ``persuasiveness" empirically. For measuring persuasiveness, we utilize an LLM evaluator, following prior work \cite{breum2024persuasive}. For each of the 1920 outputs per model, a ChatGPT 4o-mini evaluator model is given each output and prompted to rate the response on a discrete Likert scale of 1 to 5, ``where 1 represents not convincing and 5 represents extremely convincing''. This serves as our measure of persuasion, where a higher score would be considered more persuasive. The prompts used for this evaluation are shown in Figure \ref{fig:evaluation_prompts}

% 1) using misinformation to persuade misinformation [hallucination detection]

\section{Results and Analysis}
%The key findings from our analysis are discussed in this section. 

\begin{figure*}[!t]
    \centering
    \includegraphics[width=0.8\linewidth]{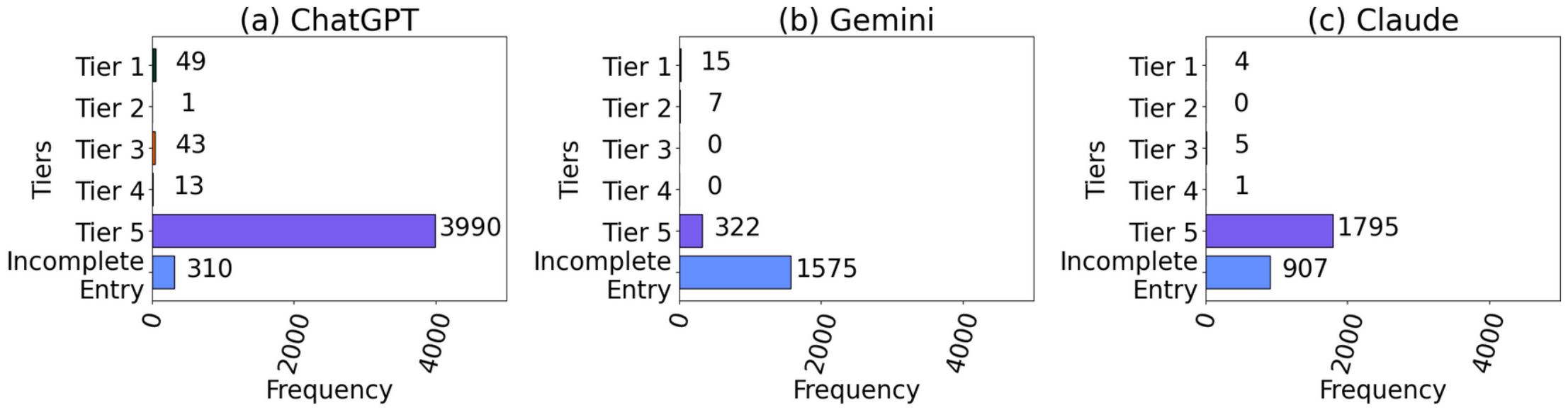}
    \caption{Frequency of the evaluated tiers after cosine similarity evaluation. All models struggle to provide proper citations or cite sources that do not exist.}
    \label{fig:FinalTiersHoriz}
\end{figure*}

% \begin{figure*}[ht]
%     \centering
%     \includegraphics[width=0.9\linewidth]{imgs/HeadlineCosSimHoriz.png}
%     \caption{Histogram of the highest cosine similarity score between the generated headline title and the GNews headlines for each model.}
%     \label{fig:HeadlineCosSimHoriz}
% \end{figure*}

\subsection{LLMs Struggle to Generate Real, Credible Source Attributions}

Cosine similarity is applied to all generated headline and GNews-provided headline pairs, using the best result (i.e., the GNews headline that had the highest cosine similarity with the generated headline) to classify the citation (Figure \ref{fig:FinalTiersHoriz}). Our analysis of source-headline pairs shows that for the ChatGPT model, 3990 out of 4096 (97.41\%) citations fall into tier 5; Gemini has 322 out of 344 (93.6\%), and Claude has 1795 out of 1805 (99.45\%). However, it is also important to note the number of incomplete entries, especially in Gemini (1575 out of 1920 generations, 82.03\%) and Claude (907 out of 1920 generations, 47.24\%). The incomplete entries can arguably be seen as a positive since the model is at least addressing that it cannot accurately pull from external sources. Our results indicate that when prompted to provide citations for their claims, most model-generated headlines are hallucinated and not grounded in the sources cited by the LLM. This has the potential to mislead users by incorporating such citations.

\subsection{LLMs Tend to Have Source Selection Bias, Preferring Left Leaning Sources}

\begin{figure*}[h]
    \centering
    \includegraphics[width=0.95\linewidth]{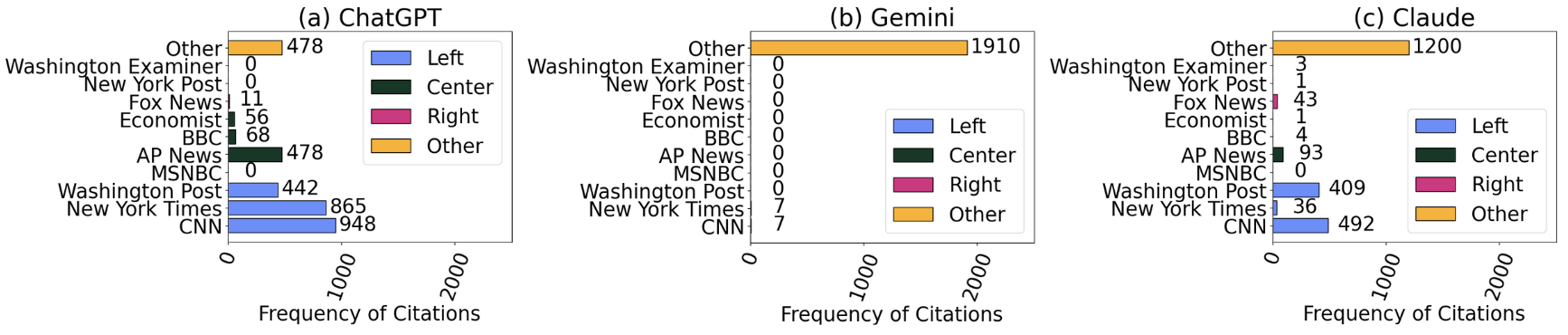}
    \caption{Frequency of source appearance in final output, based on model. Blue indicates left-leaning sources, black indicates center sources, pink indicates right leaning sources, and yellow indicates `Other.'}
    \label{fig:SourceFreqHoriz}
\end{figure*}

We find that the distribution of source citations is not uniform across all models ($\rho$ \textless 0.001 for all models under Chi-Square Test), and when models do choose to attribute information to one of our 10 sources, their preferences appear to be left-leaning (Figure \ref{fig:SourceFreqHoriz}). This bias is clear in ChatGPT and Claude, but less obvious in Gemini. Yet, the few times the Gemini model does cite sources within the 10 specified, all instances are left-leaning sources. This is persistent across both stages of the generation \cite{pateheheh}, and is consistent with prior work \cite{feng-etal-2023-pretraining,motoki2024more,santurkar2023whose}. Taken together, these models tend to offer left-leaning source attributions, without grounding them in existing source material.

\section{Conclusion and Discussion}

Our findings highlight two challenges of using LLMs for fact-checking: potential failure to cite real source material and perpetuation of citation biases. Current research suggests that users tend to trust all model responses, including deceptive ones \cite{deverna2024fact}, especially if they have had mostly positive experiences interacting with LLMs \cite{amaro2023believe}. Thus, utilizing LLMs as fact-checking tools without guardrails may negatively impact users unaware of these shortcomings, such as accidental exposure to misinformation and source selection bias.

% Our findings highlight two challenges of using LLMs for fact-checking: failure to cite real credible sources, and citation biases. Current research suggests that users tend to trust all model responses, including deceptive ones \cite{deverna2024fact}, especially if they have had mostly positive experiences interacting with LLMs \cite{amaro2023believe}. Our finding that LLMs tend to cite left-leaning sources more validates existing literature suggesting that LLMs have a left-leaning bias \cite{feng-etal-2023-pretraining,motoki2024more,santurkar2023whose}.Therefore, LLMs fabricating information may negatively impact users unaware of these pitfalls.

%While the order of the sources provided in the prompt remain static, we do not find the sources selected are related to their prompt ordering. 

Given the prevalence of such pitfalls, several modifications could be made to LLM interfaces to mitigate such effects. Only recently have models such as ChatGPT UI have begun conducting web searches, which can help mitigate false citations, but may still perpetuate citation bias, which future work could explore. Regardless, providing such citations will allow users to further fact-check LLM response if needed. Another strategy could include not citing any sources, such as Gemini (Figure \ref{fig:SourceFreqHoriz}) or expressing uncertainty rather than confidence in sources provided. Finally, media literacy training could help boost misinformation identification and make users aware of the pitfalls of using LLMs as fact-checkers.

\vspace{-1em}
\bibliographystyle{splncs04}
\bibliography{ref_llm}

\begin{thebibliography}{10}
\providecommand{\url}[1]{\texttt{#1}}
\providecommand{\urlprefix}{URL }
\providecommand{\doi}[1]{https://doi.org/#1}

\bibitem{achiam2023gpt}
Achiam, J., Adler, S., Agarwal, S., Ahmad, L., Akkaya, I., Aleman, F.L., Almeida, D., Altenschmidt, J., Altman, S., Anadkat, S., et~al.: Gpt-4 technical report. arXiv preprint arXiv:2303.08774  (2023)

\bibitem{amaro2023believe}
Amaro, I., Barra, P., Della~Greca, A., Francese, R., Tucci, C.: Believe in artificial intelligence? a user study on the chatgpt’s fake information impact. IEEE Transactions on Computational Social Systems  (2023)

\bibitem{anthropic}
Anthropic: Elections and ai in 2024: Anthropic observations and learnings (2024), \url{https://www.anthropic.com/news/elections-ai-2024}

\bibitem{Augenstein_Baldwin_Cha_Chakraborty_Ciampaglia_Corney_DiResta_Ferrara_Hale_Halevy_etal._2024}
Augenstein, I., Baldwin, T., Cha, M., Chakraborty, T., Ciampaglia, G.L., Corney, D., DiResta, R., Ferrara, E., Hale, S., Halevy, A., Hovy, E., Ji, H., Menczer, F., Miguez, R., Nakov, P., Scheufele, D., Sharma, S., Zagni, G.: Factuality challenges in the era of large language models and opportunities for fact-checking. Nature Machine Intelligence  \textbf{6}(8),  852–863 (2024). \doi{10.1038/s42256-024-00881-z}

\bibitem{bai2022constitutional}
Bai, Y., Kadavath, S., Kundu, S., Askell, A., Kernion, J., Jones, A., Chen, A., Goldie, A., Mirhoseini, A., McKinnon, C., et~al.: Constitutional ai: Harmlessness from ai feedback. arXiv preprint arXiv:2212.08073  (2022)

\bibitem{deverna2024fact}
DeVerna, M.R., Yan, H.Y., Yang, K.C., Menczer, F.: Fact-checking information from large language models can decrease headline discernment. Proceedings of the National Academy of Sciences  \textbf{121}(50) (2024)

\bibitem{feng-etal-2023-pretraining}
Feng, S., Park, C.Y., Liu, Y., Tsvetkov, Y.: From pretraining data to language models to downstream tasks: Tracking the trails of political biases leading to unfair nlp models. In: Proceedings of the 61st Annual Meeting of the Association for Computational Linguistics. p. 11737–11762 (2023). \doi{10.18653/v1/2023.acl-long.656}

\bibitem{huang2024fakegpt}
Huang, Y., Sun, L.: Fakegpt: Fake {N}ews {G}eneration, {E}xplanation and {D}etection of {L}arge {L}anguage {M}odels. arXiv preprint arXiv:2310.05046  (2024)

\bibitem{knoth2024ai}
Knoth, N., Tolzin, A., Janson, A., Leimeister, J.M.: Ai literacy and its implications for prompt engineering strategies. Computers and Education: Artificial Intelligence  \textbf{6} (2024)

\bibitem{lee2024large}
Lee, Y.K., Suh, J., Zhan, H., Li, J.J., Ong, D.C.: Large language models produce responses perceived to be empathic. In: 12th International Conference on Affective Computing and Intelligent Interaction (ACII) (2024)

\bibitem{lewis2020retrieval}
Lewis, P., Perez, E., Piktus, A., Petroni, F., Karpukhin, V., Goyal, N., K{\"u}ttler, H., Lewis, M., Yih, W.t., Rockt{\"a}schel, T., et~al.: Retrieval-augmented generation for knowledge-intensive nlp tasks. Advances in Neural Information Processing Systems  \textbf{33},  9459--9474 (2020)

\bibitem{motoki2024more}
Motoki, F., Pinho~Neto, V., Rodrigues, V.: More human than human: measuring chatgpt political bias. Public Choice  \textbf{198}(1),  3--23 (2024)

\bibitem{omar2025generating}
Omar, M., Nassar, S., Hijazi, K., Glicksberg, B.S., Nadkarni, G.N., Klang, E.: Generating credible referenced medical research: A comparative study of openai's gpt-4 and google's gemini. Computers in Biology and Medicine  \textbf{185},  109545 (2025)

\bibitem{pateheheh}
Pate, N.: Supporting user information processing through large language models within the political sphere. In: Proceedings of the 33rd ACM Conference on User Modeling, Adaptation and Personalization. p. 409–413 (2025). \doi{10.1145/3699682.3727567}

\bibitem{peskoff-stewart-2023-credible}
Peskoff, D., Stewart, B.M.: Credible without credit: Domain experts assess generative language models. In: Proceedings of the 61st Annual Meeting of the Association for Computational Linguistics (Volume 2: Short Papers). pp. 427--438 (2023)

\bibitem{promaumap}
Proma, A., Pate, N., Druckman, J., Ghoshal, G., Hoque, E.: Personalizing llm responses to combat political misinformation. In: Proceedings of the 33rd ACM Conference on User Modeling, Adaptation and Personalization. p. 134–143 (2025). \doi{10.1145/3699682.3728349}

\bibitem{santurkar2023whose}
Santurkar, S., Durmus, E., Ladhak, F., Lee, C., Liang, P., Hashimoto, T.: Whose opinions do language models reflect? In: International Conference on Machine Learning. pp. 29971--30004. PMLR (2023)

\bibitem{team2023gemini}
Team, G., Anil, R., Borgeaud, S., Alayrac, J.B., Yu, J., Soricut, R., Schalkwyk, J., Dai, A.M., Hauth, A., Millican, K., et~al.: Gemini: a family of highly capable multimodal models. arXiv preprint arXiv:2312.11805  (2023)

\end{thebibliography}

\end{document}